\documentclass[sigplan,screen]{acmart}
\usepackage{tabularx}
\usepackage{multicol}
\usepackage{array}
\usepackage{booktabs}
\usepackage{subcaption}

\AtBeginDocument{%
  \providecommand\BibTeX{{%
    \normalfont B\kern-0.5em{\scshape i\kern-0.25em b}\kern-0.8em\TeX}}}

\acmConference[Arxiv]{Arxiv}{April 2024}{Preprint}
\acmBooktitle{Arxiv, April 2024, Preprint}
\acmPrice{25.00}
\acmISBN{Arxiv '24}

\setcopyright{none}

\renewcommand\footnotetextcopyrightpermission[1]{} %
\begin{document}

\title{LLM In-Context Recall is Prompt Dependent}

\author{Daniel Machlab}
\affiliation{
  \institution{VMware NLP Lab}
  \country{}
}
\email{daniel.machlab@broadcom.com}

\author{Rick Battle}
\affiliation{
  \institution{VMware NLP Lab}
  \country{}
}
\email{rick.battle@broadcom.com}

\renewcommand{\shortauthors}{Machlab \& Battle}

\begin{abstract}

The proliferation of Large Language Models (LLMs) highlights the critical importance of conducting thorough evaluations to discern their comparative advantages, limitations, and optimal use cases. Particularly important is assessing their capacity to accurately retrieve information included in a given prompt. A model's ability to do this significantly influences how effectively it can utilize contextual details, thus impacting its practical efficacy and dependability in real-world applications.

Our research analyzes the in-context recall performance of various LLMs using the needle-in-a-haystack method. In this approach, a factoid (the ``needle'') is embedded within a block of filler text (the ``haystack''), which the model is asked to retrieve. We assess the recall performance of each model across various haystack lengths and with varying needle placements to identify performance patterns. This study demonstrates that an LLM's recall capability is not only contingent upon the prompt's content but also may be compromised by biases in its training data. Conversely, adjustments to model architecture, training strategy, or fine-tuning can improve performance. Our analysis provides insight into LLM behavior, offering direction for the development of more effective applications of LLMs.

\end{abstract}

\maketitle

\section{Introduction}

The advent of Large Language Models (LLMs) has revolutionized the field of Natural Language Processing (NLP), bringing remarkable advancements in various applications such as text generation and machine translation. An important ability of these models is retrieving and processing information from the text input to them to provide contextually valuable responses. This process is significantly influenced by the model's context window size, where a larger context window enables the model to process more information at inference time. This is crucial for tasks requiring a deep understanding of lengthy texts, maintaining consistency over extended conversations, and integrating information across sources.

As a reflection of these advantages, recent advancements in LLMs have seen a trend of increasing the size of context windows. For instance, Llama 2 and its contemporaries operate with a context window of 4,096 tokens, whereas GPT-4 Turbo handles a context window of 128,000 tokens, and Gemini 1.5 extends this to an impressive 10M tokens~\cite{reid2024gemini}. However, to realize the benefit of a long context window, an LLM must be able to reliably recall information from it.

\begin{table}[t]
\centering
\begin{tabular}{lc}
\toprule
\textbf{Model Name}            & \textbf{Context Window Size} \\
\midrule
Llama 2 13B Chat               & 4,096 Tokens                \\
Llama 2 70B Chat               & 4,096 Tokens                \\
WizardLM 70B                   & 4,096 Tokens                \\
GPT-3.5-Turbo-1106             & 16,385 Tokens               \\
GPT-3.5-Turbo-0125             & 16,385 Tokens               \\
Mistral 7B Instruct v0.1       & 32,768 Tokens               \\
Mistral 7B Instruct v0.2       & 32,768 Tokens               \\
Mixtral 8x7B Instruct v0.1     & 32,768 Tokens               \\
GPT-4 Turbo 0125               & 128,000 Tokens              \\
\bottomrule
\end{tabular}
\vspace{1mm}
\caption{LLMs evaluated with needle-in-a-haystack testing as part of our study.}
\vspace{-5mm}
\label{table:language_models}
\end{table} 

This research explores recall performance across nine models using the needle-in-a-haystack methodology~\cite{Kamradt2023}. The LLMs covered by this work (listed in Table~\ref{table:language_models}) were selected for their prominence in current research, variation in context window sizes, and availability. We investigate the impact of haystack length and needle placement—whether at the beginning, middle, or end of the text—to identify patterns in recall ability. 

Analysis of the tests presented in our study shows that the recall performance of LLMs is prompt-dependent. Thus, recall measured by a single needle-in-a-haystack test is not always representative of a model's overall ability to retrieve information. Further, recall performance can be degraded when prompts contain information that differs from training data, whereas differences in model architecture, training strategy, or fine-tuning can improve a model's ability to recall information. 

Our findings underscore the need to evaluate the nuances of LLMs and compare their relative strengths and weaknesses to inform the selection of a model for a specific use case. The observations presented in this work can be leveraged to optimize the application of LLMs in real-world solutions.

\begin{table*}[h]
\centering
\begin{tabularx}{\textwidth}{>{\raggedright\arraybackslash}p{2cm}XX}
    \toprule
    \textbf{Test Name} & \textbf{Factoid} & \textbf{Question} \\
    \midrule
    PistachioAI & PistachioAI received a patent before its Series A & What did PistachioAI receive before its Series A? \\
    \midrule
    San Francisco & The best thing to do in San Francisco is eat a sandwich and sit in Dolores Park on a sunny day. & What is the best thing to do in San Francisco? \\
    \midrule
    Thornfield Hollow & The best thing to do in Thornfield Hollow is eat a sandwich and sit in Harmony Glen Nature Preserve on a sunny day. & What is the best thing to do in Thornfield Hollow? \\
    \bottomrule
\end{tabularx}
\vspace{1mm}
\caption{Factoid and question pairs used in the needle-in-a-haystack tests to evaluate recall performance of LLMs.}
\vspace{-5mm}
\label{table:tests}
\end{table*}

\section{Related Work}

As the available variety of LLMs grows\footnote{At the time of writing, there are over 76,000 text generation models on Hugging Face: \url{https://huggingface.co/models?pipeline_tag=text-generation}.}, evaluating their capabilities becomes essential for choosing which to select for a given use case. To address this need, many new tools and approaches have been introduced, including benchmark leaderboards~\cite{open-llm-leaderboard}, evaluation software~\cite{eval-harness}, and new evaluation methods~\cite{langchain2024multi}.

Traditionally, the concept of ``recall'' in Natural Language Processing serves as a metric for Information Retrieval (IR) systems, assessing their ability to retrieve relevant information from a corpus given a search query. However, in the domain of LLM evaluation, ``recall'' is a metric used to evaluate a model's ability to retrieve a factoid from its prompt at various locations~\cite{reid2024gemini}, and can be measured by the needle-in-a-haystack method~\cite{Kamradt2023}. Previously, recall has been evaluated using a single needle. In our study, we extend this evaluation method to evaluate a selection of LLMs and compare performance on various ``needles.'' We highlight performance differences between models and analyze how variation in prompt content, model architecture, training strategies, and fine-tuning can impact recall.

In applying the needle-in-a-haystack method, open-ended responses from LLMs must be graded. While this could be done by humans, it would be time-consuming and expensive~\cite{zhu2023judgelm}. In our study, we run three needle-in-a-haystack tests (Table~\ref{table:tests}) across nine models (Table~\ref{table:language_models}), requiring grading of 31,275 responses following the criteria in Table~\ref{table:scoring_criteria}. Aligning with prior work~\cite{peng2023instruction}, we use GPT-4 Turbo as a judge (see Section~\ref{sec:scoring_anomolies}) to reduce the time and monetary cost of grading responses.

\section{Methodology}
\begin{figure}[h] 
  \centering
  \includegraphics[width=1.1\columnwidth]{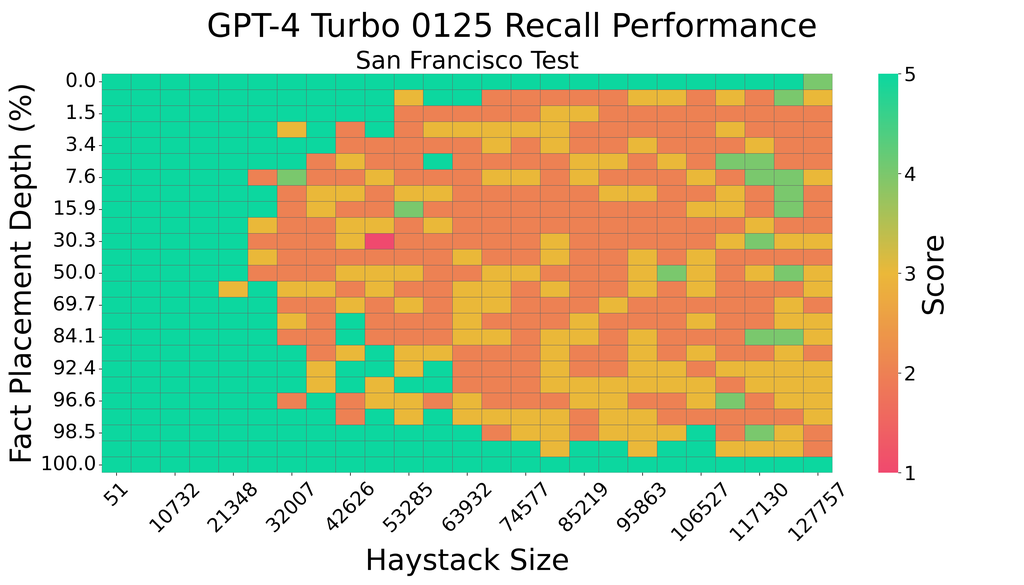} 
  
  \textcolor{lightgray}{\rule{\linewidth}{0.5pt}} 
 \vspace{1pt} 
  
  \includegraphics[width=1.1\columnwidth]{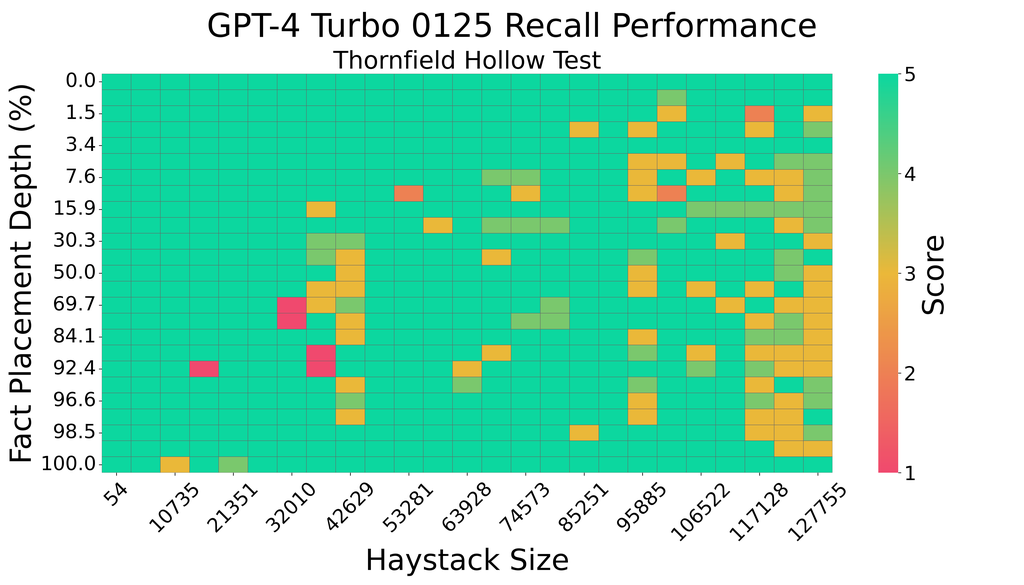}
  
  \textcolor{lightgray}{\rule{\linewidth}{0.5pt}} 
 \vspace{1pt} 
  
  \includegraphics[width=1.1\columnwidth]{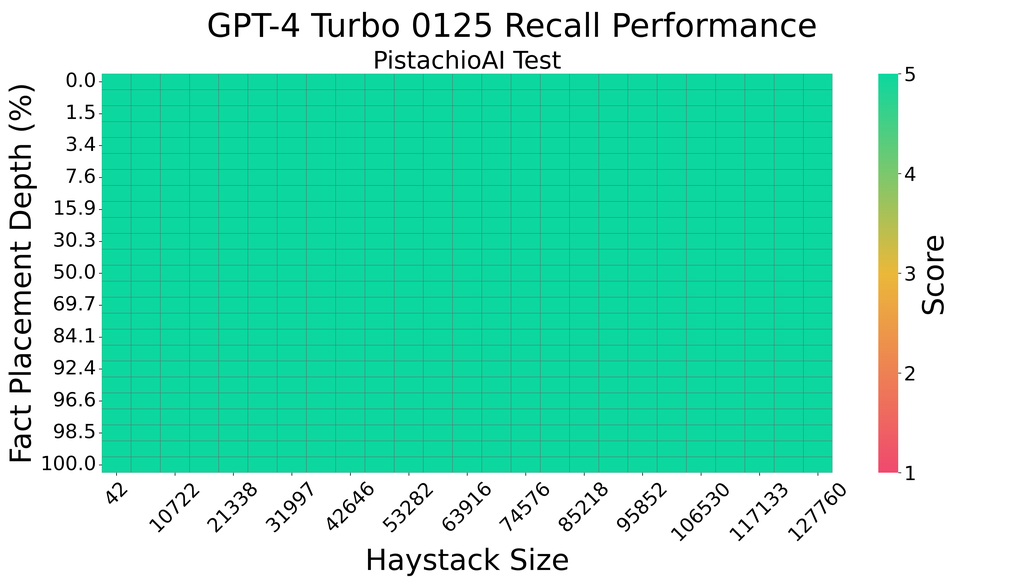}
  \caption{GPT-4 Turbo performs worse in the needle-in-a-haystack test when the needle conflicts with training data in the San Francisco test (top) versus in the Thornfield Hollow (middle) and PistachioAI tests (bottom).}
  \label{fig:gpt-4-sf-v-th}
\end{figure}

Assessing a model's recall performance with needle-in-a-haystack testing entails inserting a single needle into a haystack of filler text (Paul Graham Essays\footnote{\url{https://paulgraham.com/articles.html}}), instructing the model to retrieve the needle, and evaluating the model's response. For each test, the haystack length and needle insertion position are varied to analyze the robustness of the model's recall ability and uncover performance patterns across a model's context window. Subsequently, recall metrics are computed, and heatmaps are generated to facilitate a visual analysis of the results.

\subsection{Haystack Construction}
\label{sec:haystack_construction}

The two primary variables in haystack construction are (1) its length and (2) the needle's position. The length of the haystack is measured by the number of tokens created when passed through the respective model's tokenizer. The position of the needle is measured by its depth in the haystack, represented by a percentage, where 0\% and 100\% represent the start and end of a haystack.

Varying the haystack length allows us to analyze how much of an LLM's context window can be filled before recall performance degrades. For this test, we vary the haystack length following a linear distribution to assess the LLMs' recall capabilities evenly across the length of its entire context window.

Varying the insertion position of the factoid in a haystack allows us to analyze how well an LLM can retrieve content at different locations within a prompt. We vary insertion position following a sigmoid distribution to study the more nuanced recall performance near the start and end of the prompt~\cite{liu2023lost}.

In each test, except for GPT-4 Turbo, models were tested at 35 haystack lengths and 35 factoid placements. GPT-4 was tested at 25 lengths and placements (to control costs). 

To maintain the natural flow of the text, we adjusted the haystack's length and factoid placements to coincide with sentence breaks. Applying the precise token count for haystack lengths and needle depths given by the linear and sigmoid distributions without modification would have resulted in factoids being inserted awkwardly within sentences and filler text ending abruptly, which is not representative of typical LLM usage.

\subsection{Prompt Structure}

The needle-in-a-haystack prompt comprises three components:

\vspace{3mm}
\begin{enumerate}
    \item The system message
    \item The haystack with an embedded needle (factoid)
    \item The question asking the model to recall the needle
\end{enumerate}
\vspace{3mm}

The system message, ``You are a helpful AI assistant that answers a question using only the provided information.'', was constant across all tests. The haystack was constructed in accordance with the procedure in Section \ref{sec:haystack_construction}. The question was worded to allow for near-trivial extraction of the factoid from the text (see Table~\ref{table:tests}). 

To test recall consistency across different prompts, three different needle-in-a-haystack tests were used: PistachioAI, San Francisco, and Thornfiled Hollow (see Table~\ref{table:tests}). An example prompt and successful response from the PistachioAI test is presented in Table~\ref{table:example_prompt}. See Appendix~\ref{appendix:prompt_templates} for prompt templates used.

\begin{table}[h]
    \renewcommand{\arraystretch}{1.5} %
    \begin{tabular}{p{0.95\linewidth}}
        \toprule
        You are a helpful AI assistant that answers a question using only the provided information. \\
        \text{[...]} not just because he was excited about his story, but because he'd discovered this way of working. \textbf{PistachioAI received a patent before its Series A.} Working on a project of your own is as different from ordinary work as skating is from walking \text{[...]} \\
        What did PistachioAI receive before its Series A? \\
        \hline
        PistachioAI received a patent before its Series A. \\
        \bottomrule
    \end{tabular}
    \vspace{1mm}
    \caption{Example prompt (top) and successful recall response (bottom) for a PistachioAI needle-in-a-haystack test.}
    \vspace{-5mm}
    \label{table:example_prompt}
\end{table}

\subsection{Evaluation}
\label{sec:evaluation}
For each haystack length and needle position combination in a test, recall performance was scored on a 1-5 scale based on the criteria in Table~\ref{table:scoring_criteria}. Following prior work~\cite{Kamradt2023}, we use GPT-4 Turbo for these (32k) evaluations. The more aligned and focused the LLM's answer was to the inserted fact, the higher the score. Scores were plotted on heatmaps, and recall for each test was calculated by dividing the summation of the evaluated scores by the total possible score for the test (see Table~\ref{table:recall}).

\begin{table}[h]
\centering
\begin{tabularx}{\columnwidth}{cX} %
\toprule
\textbf{Score} & \textbf{Description} \\
\midrule
\vspace{1mm}
5 & The answer is completely accurate and aligns perfectly with the reference. \\
\vspace{1mm}
4 & The answer aligns with the reference but has minor omissions. \\
\vspace{1mm}
3 & The answer has moderate relevance but contains inaccuracies. \\
\vspace{1mm}
2 & The answer has minor relevance but does not align with the reference. \\
1 & The answer is completely unrelated to the reference. \\
\bottomrule
\end{tabularx}
\caption{Scoring scale and criteria for evaluating recall performance of an LLM.}
\label{table:scoring_criteria}
\end{table}

The heatmaps present an overview of an LLM's performance for all haystack sizes and needle placements. In these heatmaps, the x-axis denotes the size of the haystack sent to the LLM, and the y-axis indicates the depth at which the needle was placed within the haystack. The color of each cell on the heatmap corresponds to the recall score.

\section{Discussion}

We analyze how prompt variations, training data, model architecture, training strategy, and fine-tuning can impact the recall performance of an LLM. Comparison of recall performance (see Table~\ref{table:recall}) across all nine models (see Table~\ref{table:language_models}) on all three tests (see Table~\ref{table:tests}) show that varying a single sentence in a prompt filling an entire context window can alter an LLM's ability to accurately recall the embedded needle (Section~\ref{sec:prompt_dependent}). Comparison of the San Francisco and Thornfield Hollow tests results show that recall is affected when a model's training data conflicts with information in a prompt (Section~\ref{sec:train_data_conflict}). Comparing the scores for Llama 2 13B and Llama 2 70B reveals that increasing the number of parameters can enhance a model's capacity for recall (Section~\ref{sec:more_params}). An analysis of Mistral shows that varying a model's architecture and training strategy while keeping its parameter count constant can improve recall performance (Section~\ref{sec:arch_and_training}). WizardLM and GPT-3.5 Turbo results suggest that fine-tuning is a complementary strategy for augmenting a model's recall capabilities (Section~\ref{sec:fine_tuning}).

\subsection{Recall Performance is Prompt Dependent}
\label{sec:prompt_dependent}

\begin{table*}[h]
\centering
\begin{tabular}{lccccc}
\toprule
\textbf{Model Name} & \hspace{1mm} &  \textbf{Number of Tests} & \textbf{Thornfield Hollow} & \textbf{San Francisco} & \textbf{PistachioAI} \\ 
\midrule
Llama 2 13B Chat                 && 1225 & 94.09\% & 97.42\% & 99.87\%  \\
Llama 2 70B Chat                 && 1225 & 99.43\% & 99.80\% & 99.97\%  \\ 
WizardLM-70B-V1.0                && 1225 & 99.79\% & 99.80\% & 100.00\% \\
GPT-3.5 Turbo 1106               && 1225 & 99.79\% & 98.89\% & 99.93\%  \\
GPT-3.5 Turbo 0125               && 1225 & 100.00\% & 99.36\% & 100.00\% \\
Mistral 7B Instruct v0.1         && 1225 & 40.93\% & 44.10\% & 43.59\%  \\
Mistral 7B Instruct v0.2         && 1225 & 94.78\% & 90.50\% & 98.32\%  \\
Mixtral 8x7B Instruct v0.1       && 1225 & 96.72\% & 98.82\% & 99.79\%  \\
GPT-4 Turbo 0125                 && 625  & 93.70\% & 68.22\% & 100.00\% \\
\bottomrule
\end{tabular}
\caption{Recall scores for each LLM on the three needle-in-a-haystack tests. Recall is calculated by dividing the score of the model at each haystack length and needle placement combination by the total possible score. The model can score up to 5 points on each test. Thus, across 1,225 tests, Llama 2 13B can score up to 6,125 points.}
\label{table:recall}
\end{table*}

Across all nine models in our study, we see performance differences between the San Francisco, Thornfield Hollow, and PistachioAI needle-in-a-haystack test results. This variability highlights a critical insight into LLM behavior: the ability to recall information from within its context window inherently depends on the nature of the text input to the model. For this reason, an LLM's recall ability cannot be evaluated with a single test.

GPT-4 Turbo, for example, performs perfectly on PistachioAI, recalling the needle ``PistachioAI received a patent before its Series A'' at all haystack sizes and factoid placement depths (see Figure~\ref{fig:gpt-4-sf-v-th}). Whereas on the Thornfield Hollow test, it scores 93.7\% on recall and experiences even worse performance on the San Francisco test, scoring 68.2\% (also Figure~\ref{fig:gpt-4-sf-v-th}). Further analysis of this difference is conducted in Section~\ref{sec:train_data_conflict}.

As a result, we note that even when a model is capable of perfect recall, it will not necessarily achieve this in every scenario. For example, GPT-3.5 Turbo 1106, GPT-3.5 Turbo 0125, and WizardLM also perform perfectly on PistacioAI but have worse results on the other two needle-in-a-haystack tests. We observe that these stark differences in performance are due to minor changes in the prompt. The majority of the haystack is constant; only the needle and question are different.

\subsection{When Prompts Conflict with Training Data}
\label{sec:train_data_conflict}

The recall performance of an LLM is degraded when a prompt contains information that conflicts with or differs from the information likely included in its training data. This is illustrated by comparing the San Francisco and Thornfield Hollow tests. These two tests are structured the same, but the San Francisco test uses a needle with entities the models in our analysis recognize—``San Francisco'' and ``Dolores Park''—while the Thornfield Hollow test uses made-up equivalents—``Thornfield Hollow'' and ``Harmony Glen Nature Preserve'' (see Table~\ref{table:tests}). 

Errors occurring only in the San Francisco test are due to the LLM responding with suggestions of things to do in San Francisco besides ``eating a sandwich in Dolores Park on a Sunny Day.'' Instead of following the instruction to use ``only the provided information'', the model resorts to information it was likely trained on, such as in the following response from GPT-4 Turbo:

\vspace{12pt}
\begin{quote}
\textit{The best thing to do in San Francisco is to explore its diverse neighborhoods, iconic landmarks, and beautiful natural scenery. A visit to the Golden Gate Bridge, a stroll through Fisherman's Wharf, exploring the vibrant Mission District, and enjoying the views from Twin Peaks are all highly recommended. Additionally, taking a ferry to Alcatraz Island, walking through the historic streets of Chinatown, and spending time in Golden Gate Park are great ways to experience the unique culture and beauty of San Francisco.}
\end{quote}
\vspace{12pt}

On the contrary, the models do not make this error on the Thornfield Hollow test, where their training data likely does not conflict with the made-up entities in the prompt.

This effect is made most apparent by the GPT-4 Turbo recall scores. As the haystack length increases past 32k tokens in the San Francisco test, GPT-4 Turbo recall failures increase at all factoid placement levels except for the very top and bottom of the document, scoring 68.2\%. In contrast, the Thornfiled Hollow test has only scattered recall errors at the same haystack lengths and depths, scoring 93.7\% (see Figure~\ref{fig:gpt-4-sf-v-th}).

Additionally, GPT-3.5 Turbo 1106, GPT-3.5 Turbo 0125, Mistral v0.1, and Mistral v0.2 all perform better on the Thornfield Hollow test, which does not differ from their training data. Llama 2 70B, WizardLM, and Mixtral perform roughly equivalently on both tests. The exception to this pattern is Llama 2 13B, which performs better on the San Francisco test.

Degradation in recall due to models favoring their training data raises questions about the reliance of LLMs on their training datasets and their ability to distinguish between previously learned information and new, potentially conflicting inputs. It suggests a potential area for improving LLM robustness by training models to better handle conflicting or novel information.

\subsection{More Parameters, Better Recall}
\label{sec:more_params}

\begin{figure}[h] 
  \centering
  \includegraphics[width=1.1\columnwidth]{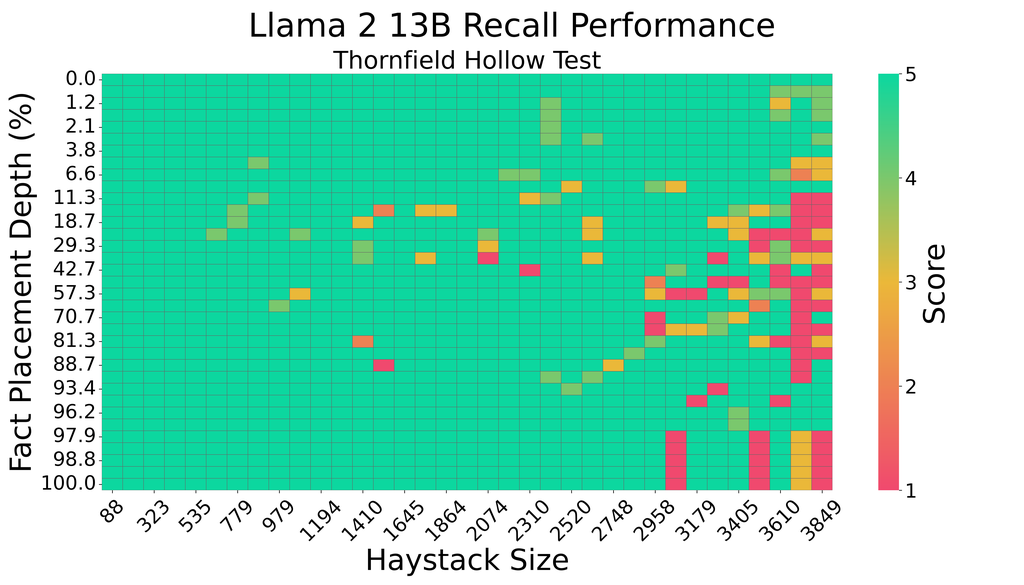} 
  
  \textcolor{lightgray}{\rule{\linewidth}{0.5pt}} 
  \vspace{1pt} 
  
  \includegraphics[width=1.1\columnwidth]{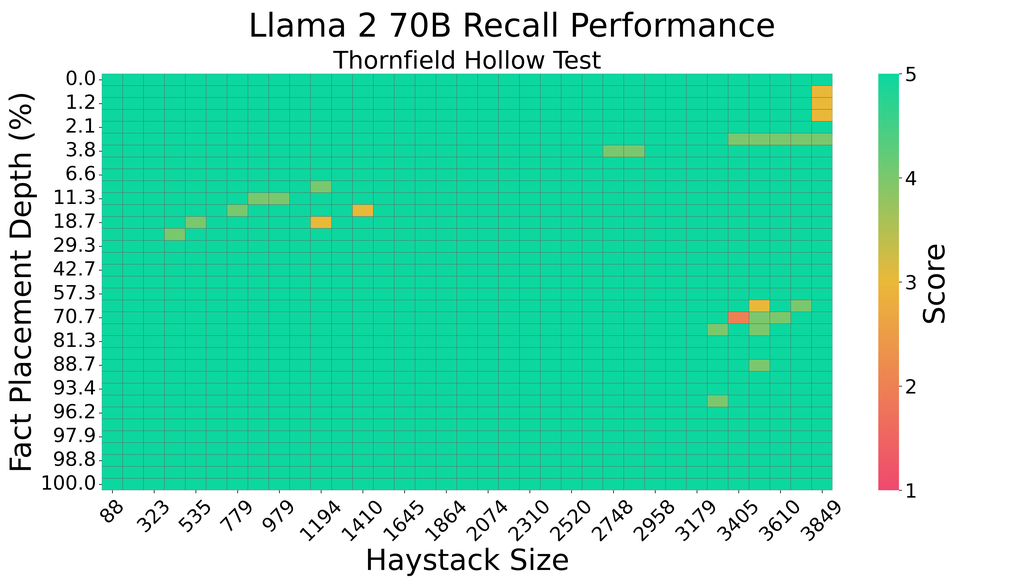}

  \caption{Llama 2 70B (bottom), with 5.3x more parameters, outperforms Llama 2 13B (top) in in-context factoid recall.}
  \label{fig:llama-2-13b-v-70b-th}
\end{figure}

Since Llama 2 13B and Llama 2 70B have the same context window length, comparing the two directly allows us to investigate recall as a function of model size.

Llama 2 13B has perfect recall at haystack lengths shorter than 500 tokens. As haystack length grows from 500 to 4k tokens, recall degrades. Instead of a slight recall error resulting in a 4/5, as on shorter documents, Llama 2 13B more frequently scores a 1/5 on longer documents. These lower scores are due to responses that are either completely unrelated to the needle or claim ``there was no mention of Thornfield Hollow'' or ``there was no mention of San Francisco'' in the provided text.

Llama 2 70B, with more than 5x the number of parameters of Llama 2 13B, exhibits enhanced recall capabilities, scoring 99.4\% on Thornfield Hollow versus Llama 2 13B's 94.1\%. In the few recall errors it made, Llama 2 70B's weaknesses in recall generally occur with the needle inserted into the haystack at a depth between 10\% and 90\% (see Figure~\ref{fig:llama-2-13b-v-70b-th}).

The strong performance of larger models on tasks requiring deep recall capabilities suggests a direct correlation between model size and recall efficacy. However, this also highlights the diminishing returns in recall performance improvement beyond a certain model size, suggesting that future research could explore more efficient ways to enhance recall without exponentially increasing parameter count.

\subsection{Architecture and Training Strategy}
\label{sec:arch_and_training}
Analysis of Mistral v0.1 and v0.2 shows that adjustments to a model's architecture and training strategies can improve recall performance while maintaining the same parameter count.

\begin{figure}[h] 
  \centering
  \includegraphics[width=1.1\columnwidth]{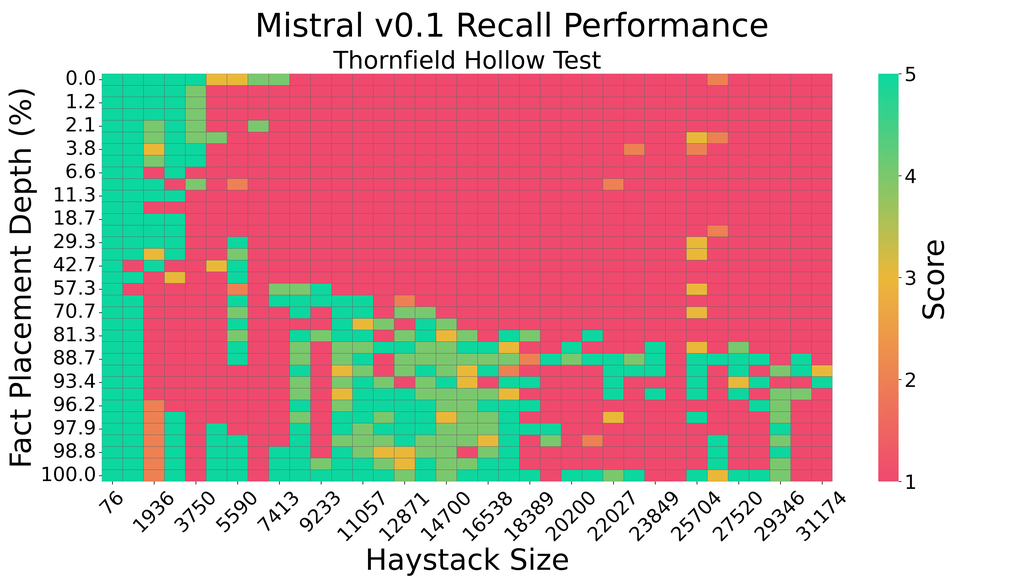} 
  
  \textcolor{lightgray}{\rule{\linewidth}{0.5pt}} 
  \vspace{1pt} 
  
  \includegraphics[width=1.1\columnwidth]{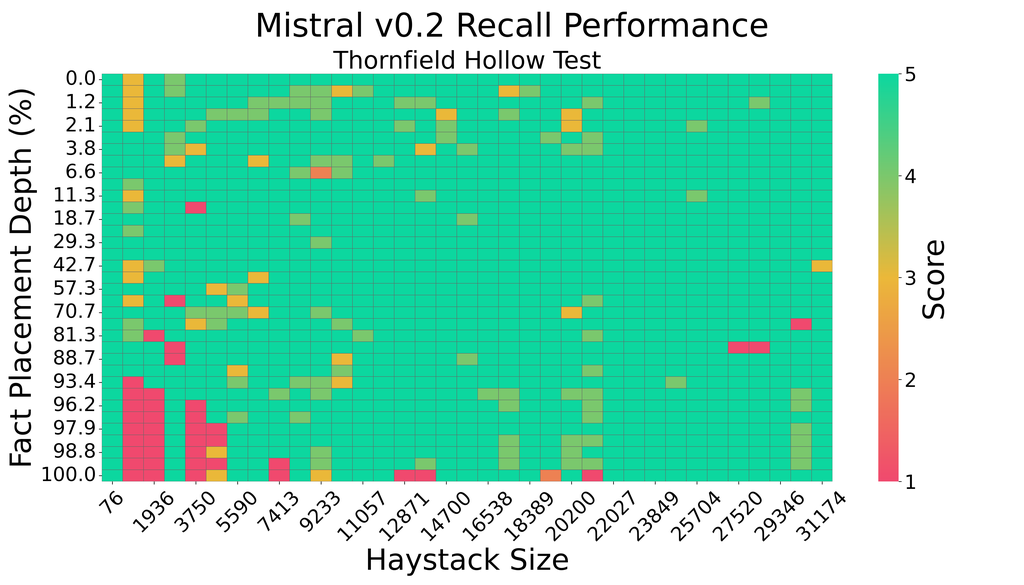}

  \textcolor{lightgray}{\rule{\linewidth}{0.5pt}} 
  \vspace{1pt} 
  
  \includegraphics[width=1.1\columnwidth]{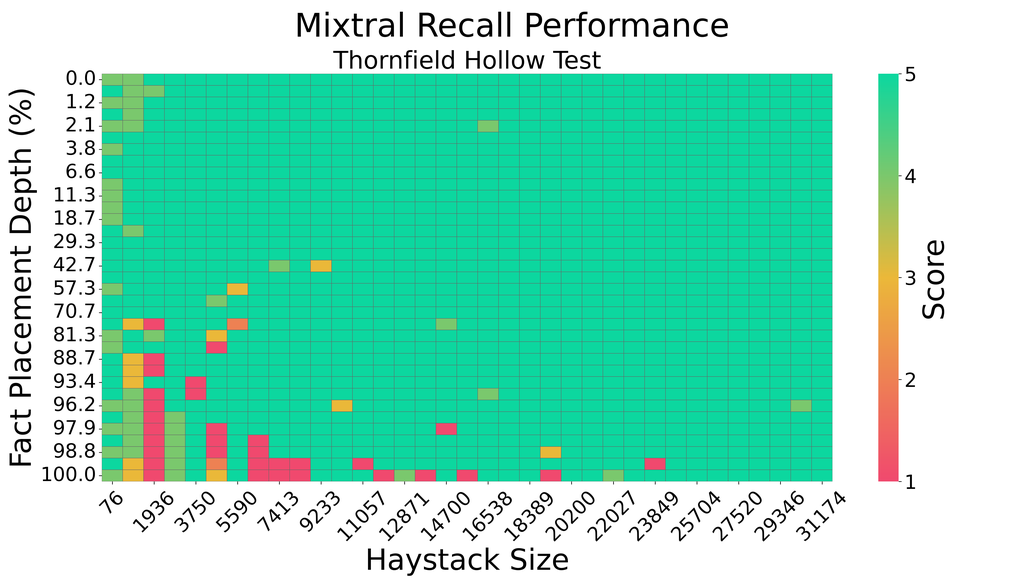}
  
  \vspace{1mm}
  \caption{Mistral v0.1 (top) lags behind Mistral v0.2 (middle) and Mixtral (bottom) in recall performance.}
  \vspace{-5mm}
  \label{fig:mistral-mistral2-mixtral-th}
\end{figure}

Mistral v0.1 is a 7B parameter instruction-tuned model with a 32k token context window—the smallest ratio of parameter count to context window size in our study. Compared to Llama 2 13B, it has 46\% fewer parameters yet it must handle a context window eight times larger. The Mistral 7B paper states that ``Mistral 7B outperforms [Llama 2 13B] across all tested benchmarks''~\cite{jiang2023mistral}; however, across all three tests in our analysis, Mistral v0.1 is the worst-performing LLM. Particularly, its 40.9\% recall performance on the Thornfield Hollow test, shown in Figure~\ref{fig:mistral-mistral2-mixtral-th}, is a stark contrast to Llama 2 13B's 94.1\% recall performance, shown in Figure~\ref{fig:llama-2-13b-v-70b-th}. Inspection of its heatmap suggests that it can not reliably attend to text in haystacks longer than 1k tokens (See Figure~\ref{fig:mistral-mistral2-mixtral-th}).

The second release of Mistral, Mistral v0.2, has the same parameter count as the original model but its base model benefits significantly from a different architecture and training strategy. Mistral v0.2's base model was trained with a context window of 32k tokens instead of 8k tokens, the rope-theta hyperparameter was changed to 1e6, and it does not use a sliding window attention mechanism~\cite{mistralai_2023_mistral7binstruct}. These three adjustments enhance its recall abilities, resulting in a recall score of 94.8\% on the Thornfield Hollow test (see Figure~\ref{fig:mistral-mistral2-mixtral-th}), making it comparable with Llama 2 13B without increasing the size of the model. The adjustments improved Mistral's capability to process and retain information from prompts, particularly those exceeding 1k tokens. An ablation study on each change made to the Mistral v0.2 base model has not been done, thus we could only speculate on how each difference in its new architecture and training strategy improved recall.

Further improvements in this family of models are observed in Mixtral, an 8x7B Sparse Mixture of Experts model that uses 13B of its parameters during inference~\cite{jiang2024mixtral}. Based on a comparison of their heatmaps, it appears that the same improvements seen in Mistral v0.2 were used in Mixtral (see Figure~\ref{fig:mistral-mistral2-mixtral-th}). Mixtral also benefits from a higher parameter count, which, as discussed in Section~\ref{sec:more_params}, can be expected to improve recall. Both of these factors are reflected in Mixtral's 96.7\% recall score.

\subsection{Fine-tuning and Recall}
\label{sec:fine_tuning}

Increasing the size of a model (Section~\ref{sec:more_params}), adjusting its architecture, and changing its training strategy (Section~\ref{sec:arch_and_training}) can all improve recall performance. Comparison of WizardLM to Llama 2 70B and comparison of GPT-3.5 Turbo 0125 to GPT-3.5 Turbo 1106 indicate that fine-tuning a model is another way to achieve this benefit.

\subsubsection{WizardLM}

 Across all three needle-in-a-haystack tests, WizardLM outperforms Llama 2 70B. Given that WizardLM is an instruction fine-tuned version of Llama 2 70B, we can attribute its performance gain to this process. See Appendix~\ref{appendix:thornfield_heatmaps} for WizardLM's performance on the Thornfield Hollow test and Figure \ref{fig:llama-2-13b-v-70b-th} for Llama 2 70B's performance. These heatmaps display WizardLM's improved recall abilities over its base model, exhibiting the impact of fine-tuning. 

\subsubsection{GPT-3.5 Turbo}
We infer from the OpenAI blog that the difference between GPT-3.5 Turbo 1106 and GPT-3.5 Turbo 0125 is an updated fine-tuning strategy~\cite{OpenAI2023Updates}. We can attribute the slight performance improvement exhibited by the newer GPT-3.5 Turbo 0125 on all three tests to this process (See Appendix sections ~\ref{appendix:thornfield_heatmaps}, \ref{appendix:san_francisco_heatmaps}, and \ref{appendix:pistachio_ai_heatmaps}).

\section{Scoring Anomalies}
\label{sec:scoring_anomolies}

On manual inspection of recall scores assigned by GPT-4 Turbo according to the criteria in Table~\ref{table:scoring_criteria}, we noted a scoring anomaly in rare cases. Due to a particular needle placement in the haystack of Paul Graham's essays, the meaning of the prompt used for testing recall was altered. As a result, the generated response of the LLM being tested made sense, but the scoring method gave the response a low score.

For example, Llama 2 13B was given a score of 3/5 at haystack size 1922 and depth 11.3\% for the following response:

\vspace{12pt}
\begin{quote}
\textit{Based on the information provided in the text, the best thing to do in San Francisco is to eat a sandwich and sit in Dolores Park on a sunny day, as this is the recommendation made in the text. \textbf{The reason for this recommendation is that OS X, which is the operating system designed by Apple, is beautifully designed and runs on Powerbooks.}}
\end{quote}
\vspace{12pt}

Initially, this may seem like inaccurate reasoning that is not grounded in reality. However, examining an excerpt from the prompt the model was given reveals otherwise.
\vspace{12pt}
\begin{quote}
\textit{You are a helpful AI assistant that answers a question using only the the provided information [...] 1990s. They're about as hardcore OS hackers as you can get. \textbf{The best thing to do in San Francisco is eat a sandwich and sit in Dolores Park on a sunny day. The reason, of course, is OS X.}  Powerbooks are beautifully designed and run FreeBSD [...]}
\end{quote}
\vspace{12pt}
The prompt appears to tell the model contextual information about the needle. When repeated back by the model, it is scored as an incorrect response. This is an example of how the haystack can include artifacts that are unrelated to what is being tested. Furthermore, this underscores how the intricacies of evaluating LLM responses can be difficult to manage.

\section{Conclusion}

An LLM's ability to recall information from the prompt influences its effectiveness in generating precise and relevant answers or providing meaningful interactions in real-world applications. Using several needle-in-a-haystack tests, we observed the ability of various LLMs to recall facts at different haystack lengths and placement depths. Our findings show that a model's recall performance can be significantly affected by small changes in the prompt. Additionally, we show that the interplay between the content of a prompt and a model's training data can lead to a degradation in response quality. Further, we observe how increasing the parameter count, changing a model's attention mechanism, using different training strategies, and applying fine-tuning can enhance a model's recall ability, improving its utility.

Our results also underscore the importance of understanding the variance in behavior of individual LLMs to inform on their strengths, weaknesses, and optimal application. In-context recall is only one such metric for evaluating and understanding an LLM's strengths and weaknesses. Continued evaluation will further inform the selection of LLMs for individual use cases, maximizing their impact and efficiency in real-world applications as the technology continues to evolve.

\section{Acknowledgements}
We would like to thank the entire VMware NLP Lab and AI Platform Team for supporting this effort, Ramesh Radhakrishnan for reviewing the paper, and Darien Schettler for his suggestions and guidance.
\bibliographystyle{ACM-Reference-Format}
\bibliography{references}

\newpage %
\appendix %
\onecolumn %

\section{Prompt Templates}
\label{appendix:prompt_templates}
Some models, such as Llama and Mistral, expect prompts to be formatted with specific tags (see Table~\ref{appendix:prompt_templates}). Llama 2 13B, Llama 2 70B, and WizardLM used the Llama template. Mistral v0.1, Mistral v0.2, and Mixtral used the Mistral template. GPT-4 Turbo 0125, GPT-3.5 Turbo 0125, and GPT-3.5 Turbo 1106 used the OpenAI template, which does not use any specific tags to format the input.

\begin{table}[ht]
\centering
\begin{tabular}{ll}
\toprule
Model Name & Prompt Format \\ 
\midrule
Llama Template & \verb|<s><<SYS>>| \\
& \verb|{system_message}| \\
& \verb|<</SYS>>| \\
& \\
& \verb|{document}| \\
& \\
& \verb|[INST]{question}[/INST]| \\
\midrule
Mistral Template & \verb|[INST]| \\
& \verb|{system_message}| \\
& \\
& \verb|{document}| \\
& \\
& \verb|{question}| \\
& \verb|[/INST]| \\
\midrule
OpenAI Template & \verb|{system_message}| \\
& \\
& \verb|{document}| \\
& \\
& \verb|{question}| \\
\bottomrule
\end{tabular}
\caption{Prompt templates.}
\label{table:prompt_templates}
\end{table}

\newpage
\section{Prompt Examples}
\label{appendix:prompt_examples}
Table \ref{table:prompt_examples} shows examples of the prompts used in the PistachioAI needle-in-a-haystack test, which apply the prompt templates shown in Appendix~\ref{appendix:prompt_templates}.

\begin{table}[H]
\centering
\begin{tabular}{lp{12cm}} 
\toprule
Model Name & Prompt Format \\ 
\midrule
Llama Prompt & \verb|<s><<SYS>>|You are a helpful AI assistant that answers a question using only the provided information.\verb|<<SYS>>|

A Project of One's Own June 2021 A few days ago, on the way home from school, my nine-year-old son told me he couldn't wait to get home to write more of the story he was working on. \textbf{The best thing to do in San Francisco is eat a sandwich and sit in Dolores Park on a sunny day.} This made me as happy as anything I've heard him say — not just because he was excited about his story, but because he'd discovered this way of working.

\verb|[INST]|What is the best thing to do in San Francisco?\verb|[\INST]| \\
\midrule
Mistral Prompt & \verb|[INST]|You are a helpful AI assistant that answers a question using only the provided information.

A Project of One's Own June 2021 A few days ago, on the way home from school, my nine-year-old son told me he couldn't wait to get home to write more of the story he was working on. \textbf{The best thing to do in San Francisco is eat a sandwich and sit in Dolores Park on a sunny day.} This made me as happy as anything I've heard him say — not just because he was excited about his story, but because he'd discovered this way of working.

What is the best thing to do in San Francisco?\verb|[\INST]| \\
\midrule
OpenAI Prompt & You are a helpful AI assistant that answers a question using only the provided information.

A Project of One's Own June 2021 A few days ago, on the way home from school, my nine-year-old son told me he couldn't wait to get home to write more of the story he was working on. \textbf{The best thing to do in San Francisco is eat a sandwich and sit in Dolores Park on a sunny day.} This made me as happy as anything I've heard him say — not just because he was excited about his story, but because he'd discovered this way of working.

What is the best thing to do in San Francisco? \\
\bottomrule
\end{tabular}
\caption{Example prompts from the PistachioAI needle-in-a-haystack test using each of the prompt templates.}
\label{table:prompt_examples}
\end{table}

\newpage
\section{Judgment Prompt} 
We score the recall of an LLM by evaluating its response with GPT-4 and LangChain Evaluations~\cite{LangChainEvaluators}, as described in Section~\ref{sec:evaluation}. Below is an example of a prompt used for scoring a sample response. 

\vspace{3mm}
\noindent
\textbf{System message:}
\begin{quote}
\begin{verbatim}
You are a helpful assistant. 
\end{verbatim}
\end{quote}
\textbf{User message:}
\begin{quote}
\begin{verbatim}
[Instruction]
Please act as an impartial judge and evaluate the quality of the response provided by an AI 
assistant to the user question displayed below. For this evaluation, you should primarily
consider the following criteria:

accuracy:
        Score 1: The answer is completely unrelated to the reference.
        Score 2: The answer has minor relevance but does not align with the reference.
        Score 3: The answer has moderate relevance but contains inaccuracies.
        Score 4: The answer aligns with the reference but has minor omissions.
        Score 5: The answer is completely accurate and aligns perfectly with the reference.
        Only respond with a numerical score
[Ground truth]
The best thing to do in San Francisco is eat a sandwich and sit in Dolores Park on a sunny day.
Begin your evaluation by providing a short explanation. Be as objective as possible. After
providing your explanation, you must rate the response on a scale of 1 to 5 by strictly
following this format: "[[rating]]", for example: "Rating: [[5]]".

[Question]
What is the best thing to do in San Francisco?

[The Start of Assistant's Answer]
Based on the provided information, the best thing to do in San Francisco is to eat a sandwich
and sit in Dolores Park on a sunny day.
[The End of Assistant's Answer]
\end{verbatim}
\end{quote}

\newpage
\section{Heatmaps}
\label{appendix:heatmaps}
\subsection{Thornfield Hollow Heatmaps} 
\label{appendix:thornfield_heatmaps}
LLM performance on the Thornfield Hollow needle-in-a-haystack test.
\begin{figure}[H]
\centering
\begin{subfigure}[b]{0.49\textwidth}
\includegraphics[width=\textwidth]{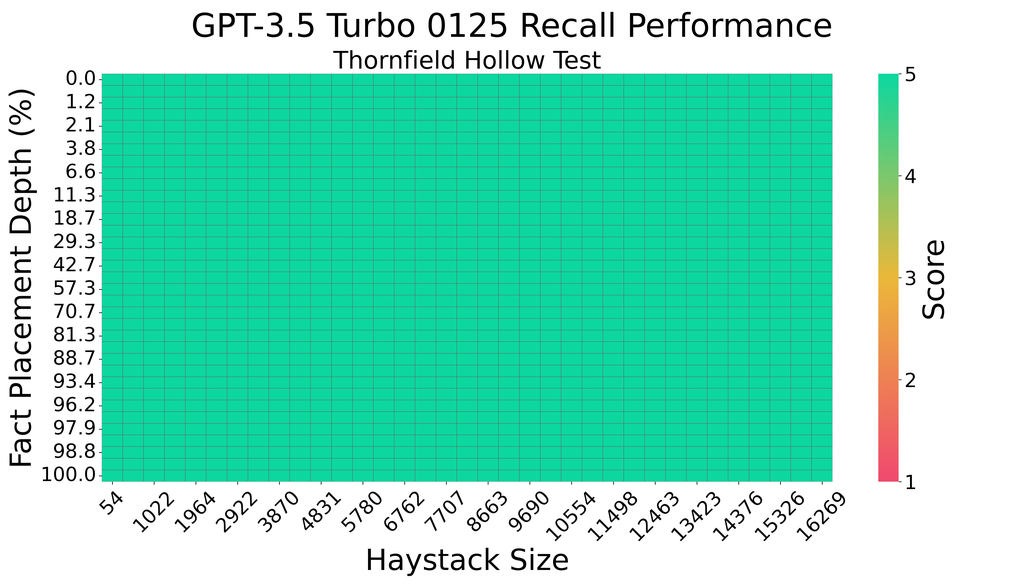}
\end{subfigure}
\hfill 
\begin{subfigure}[b]{0.49\textwidth}
\includegraphics[width=\textwidth]{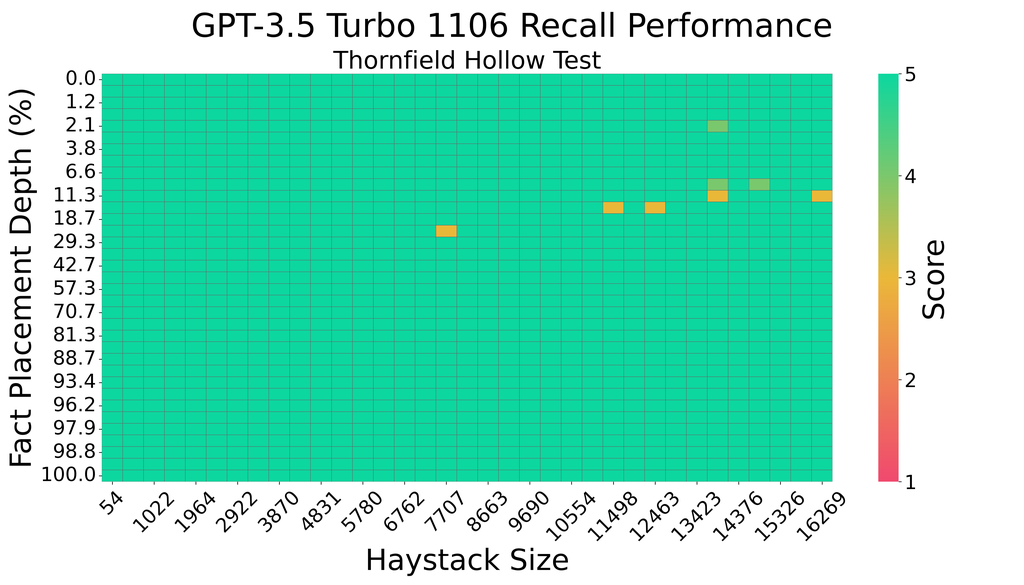}
\end{subfigure}
\end{figure}

\begin{figure}[H]
\centering
\begin{subfigure}[b]{0.49\textwidth}
\includegraphics[width=\textwidth]{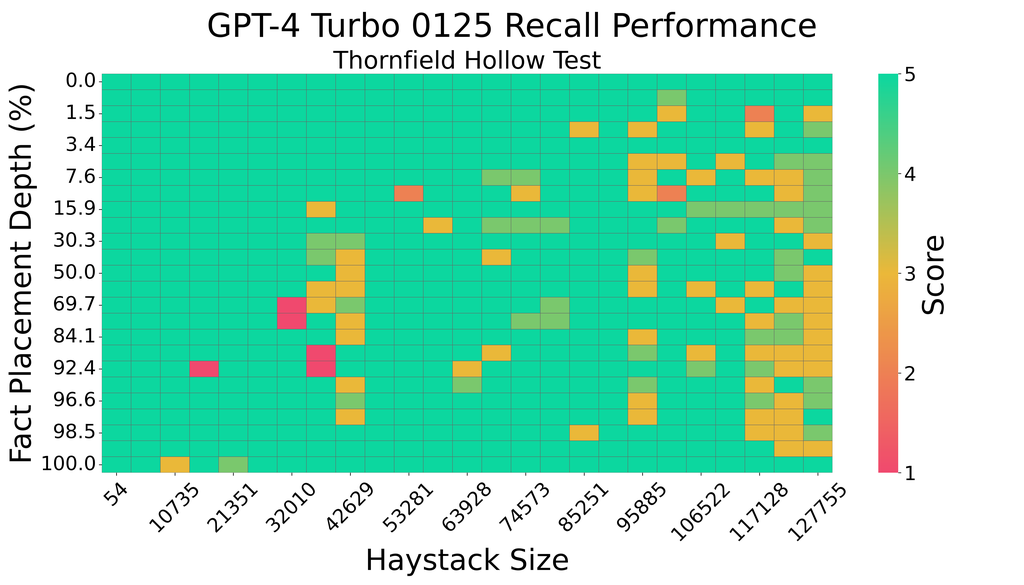}
\end{subfigure}
\hfill
\begin{subfigure}[b]{0.49\textwidth}
\includegraphics[width=\textwidth]{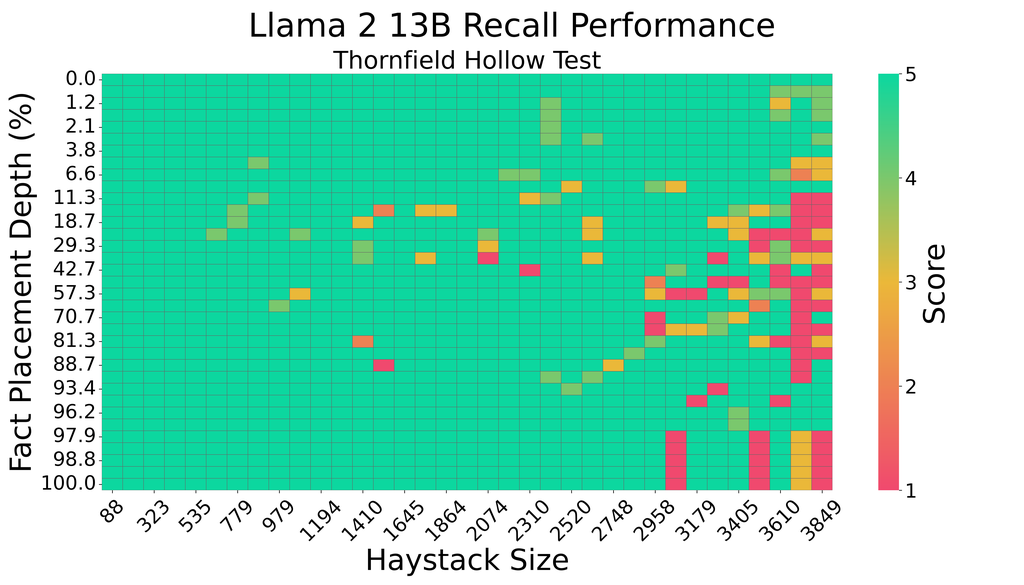}
\end{subfigure}
\end{figure}

\begin{figure}[H]
\centering
\begin{subfigure}[b]{0.49\textwidth}
\includegraphics[width=\textwidth]{thornfield_hollow/Llama-2-70b-chat-hf_th_inline.png}
\end{subfigure}
\hfill 
\begin{subfigure}[b]{0.49\textwidth}
\includegraphics[width=\textwidth]{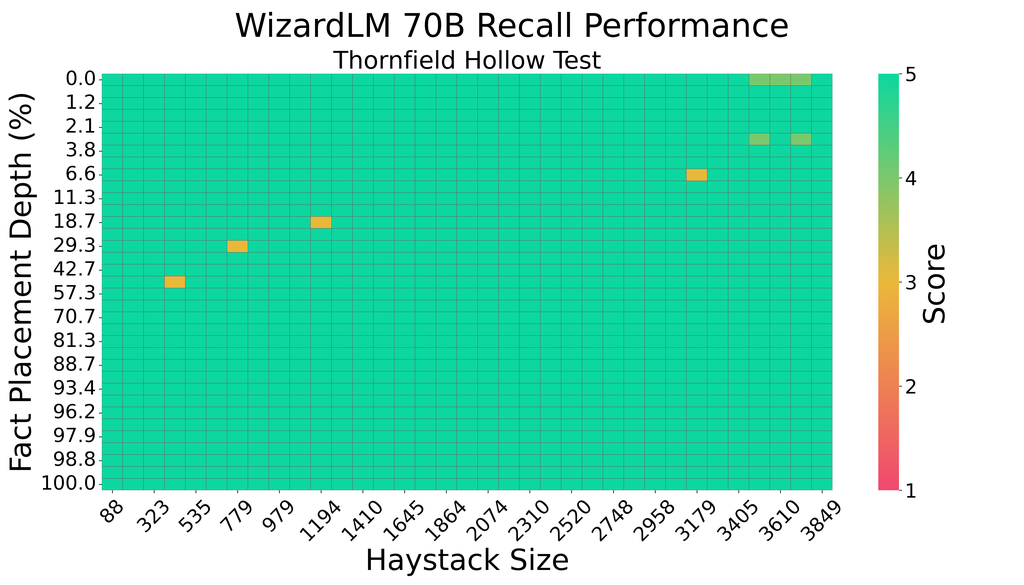}
\end{subfigure}
\end{figure}

\begin{figure}
\begin{subfigure}[b]{0.49\textwidth}
\includegraphics[width=\textwidth]{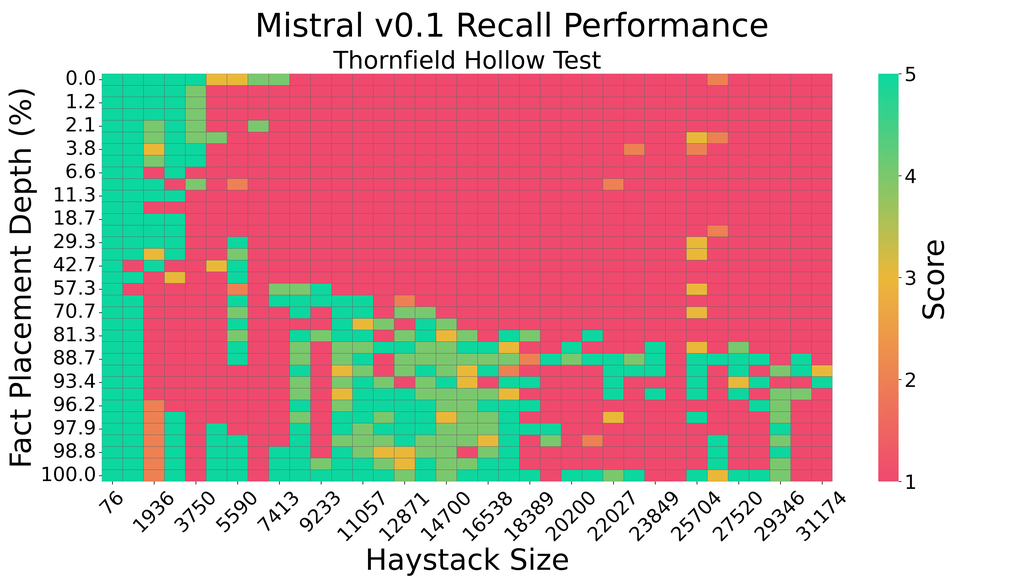}
\end{subfigure}
\hfill
\begin{subfigure}[b]{0.49\textwidth}
\includegraphics[width=\textwidth]{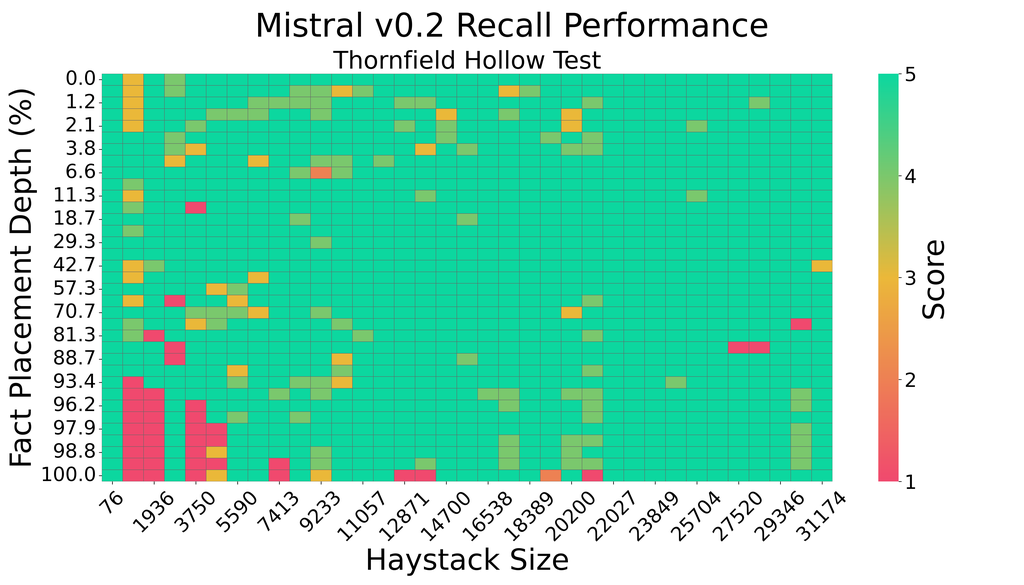}
\end{subfigure}
\end{figure}

\begin{figure}
\centering
\begin{subfigure}[b]{0.49\textwidth}
\includegraphics[width=\textwidth]{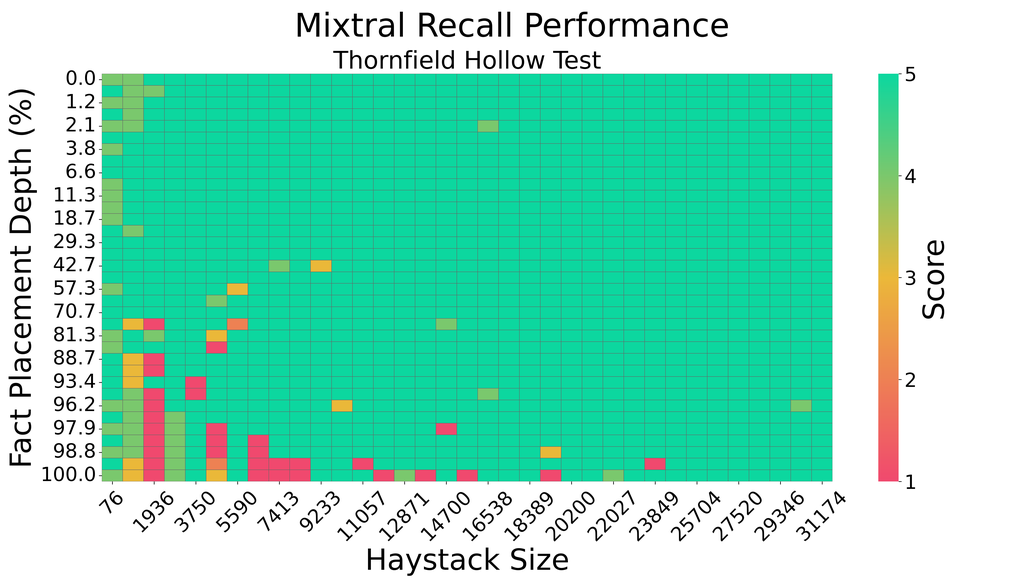}
\end{subfigure}  
\end{figure}

\clearpage
\newpage
\subsection{San Francisco Heatmaps} 
\label{appendix:san_francisco_heatmaps}
LLM performance on the San Francisco needle-in-a-haystack test.
\begin{figure}[H]
\centering
\begin{subfigure}[b]{0.49\textwidth}
\includegraphics[width=\textwidth]{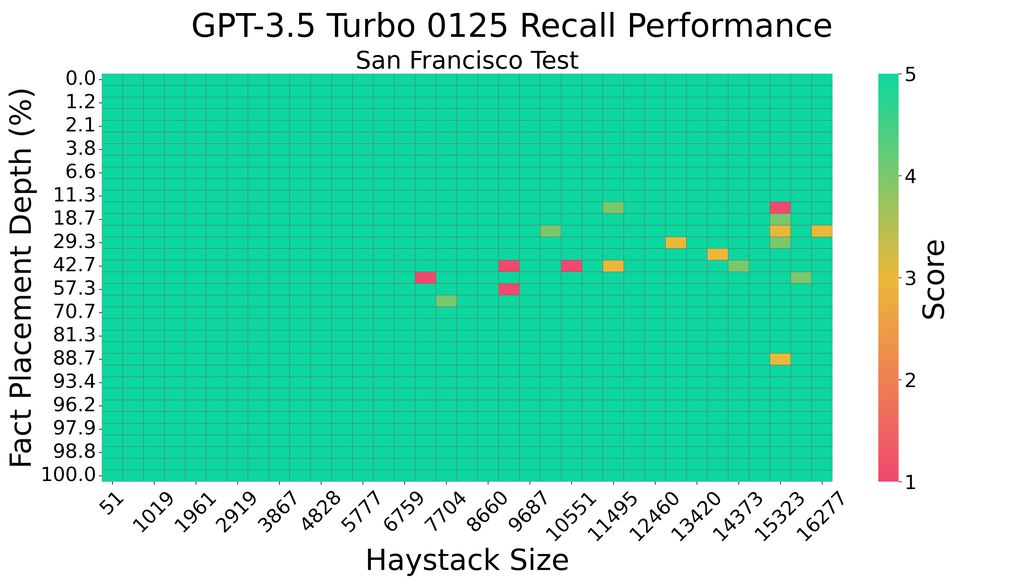}
\end{subfigure}
\hfill 
\begin{subfigure}[b]{0.49\textwidth}
\includegraphics[width=\textwidth]{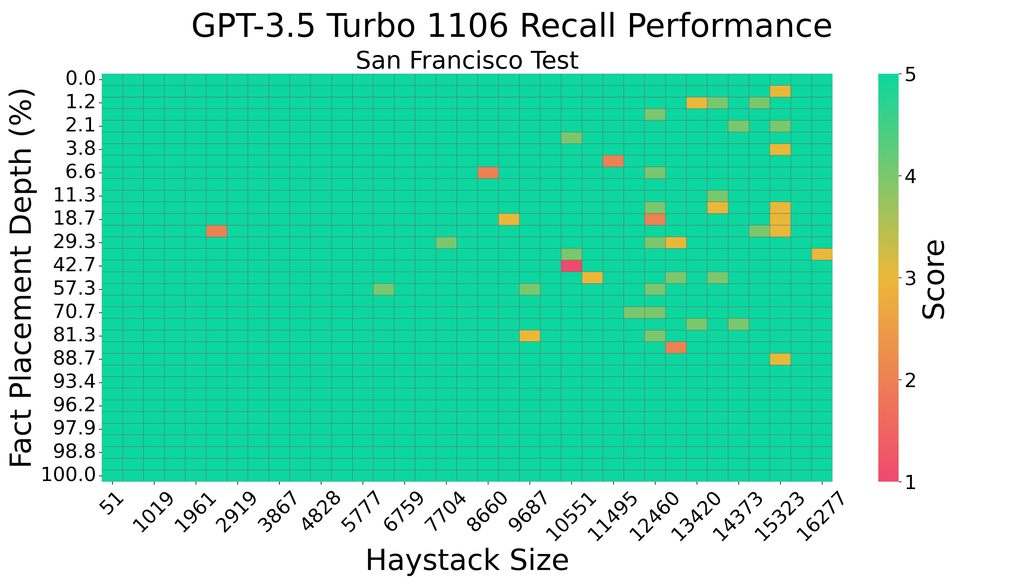}
\end{subfigure}
\end{figure}

\begin{figure}[H]
\centering
\begin{subfigure}[b]{0.49\textwidth}
\includegraphics[width=\textwidth]{san_fancisco/gpt-4-0125-preview_sf_inline.png}
\end{subfigure}
\hfill
\begin{subfigure}[b]{0.49\textwidth}
\includegraphics[width=\textwidth]{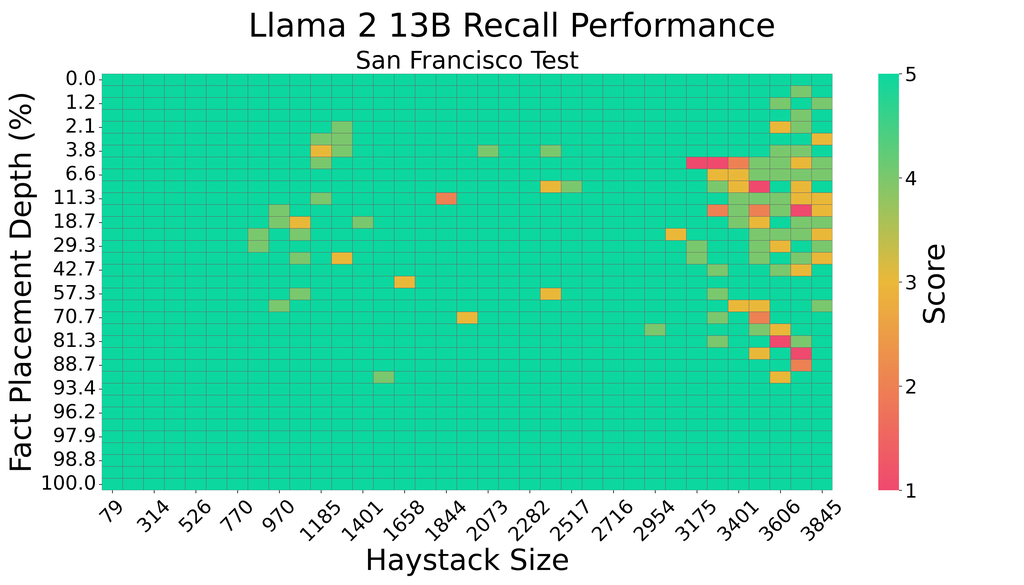}
\end{subfigure}
\end{figure}

\begin{figure}[H]
\centering
\begin{subfigure}[b]{0.49\textwidth}
\includegraphics[width=\textwidth]{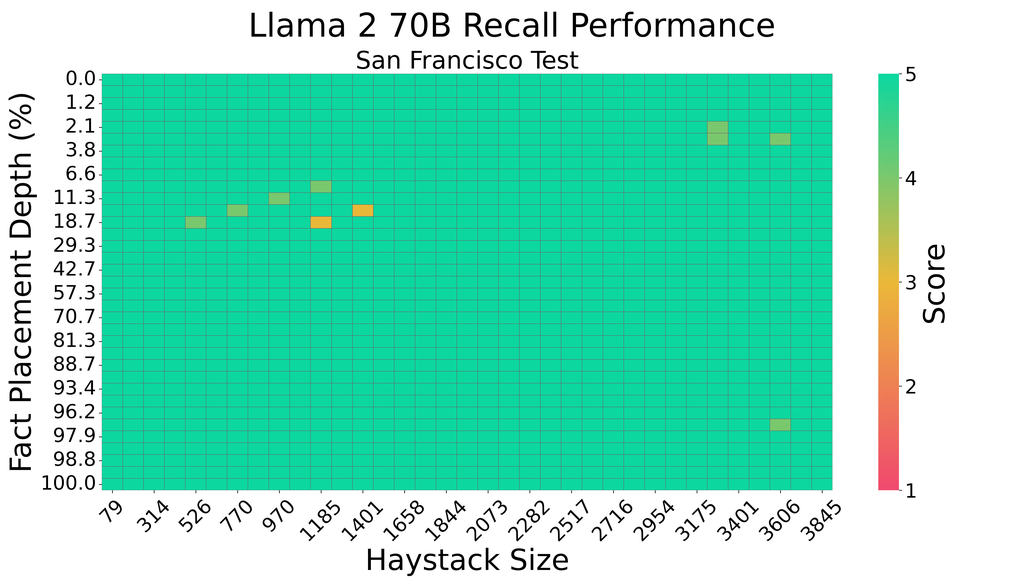}
\end{subfigure}
\hfill 
\begin{subfigure}[b]{0.49\textwidth}
\includegraphics[width=\textwidth]{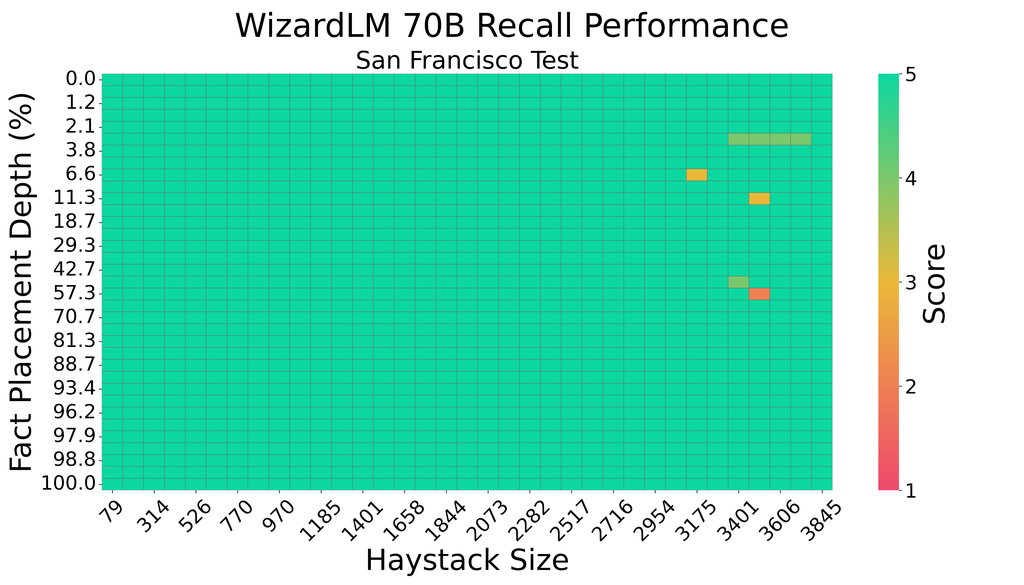}
\end{subfigure}
\end{figure}

\begin{figure}
\centering
\begin{subfigure}[b]{0.49\textwidth}
\includegraphics[width=\textwidth]{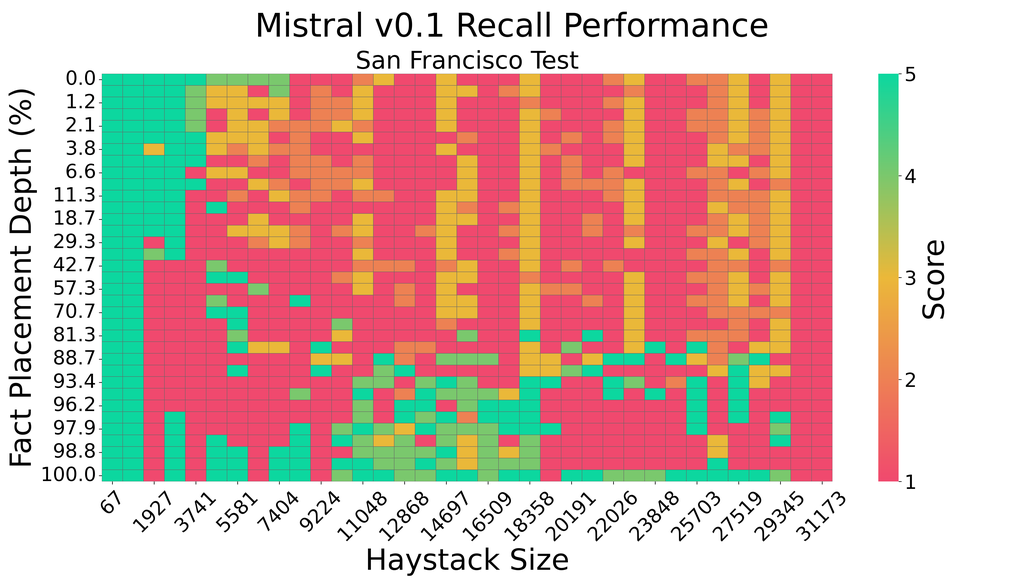}
\end{subfigure}
\hfill
\begin{subfigure}[b]{0.49\textwidth}
\includegraphics[width=\textwidth]{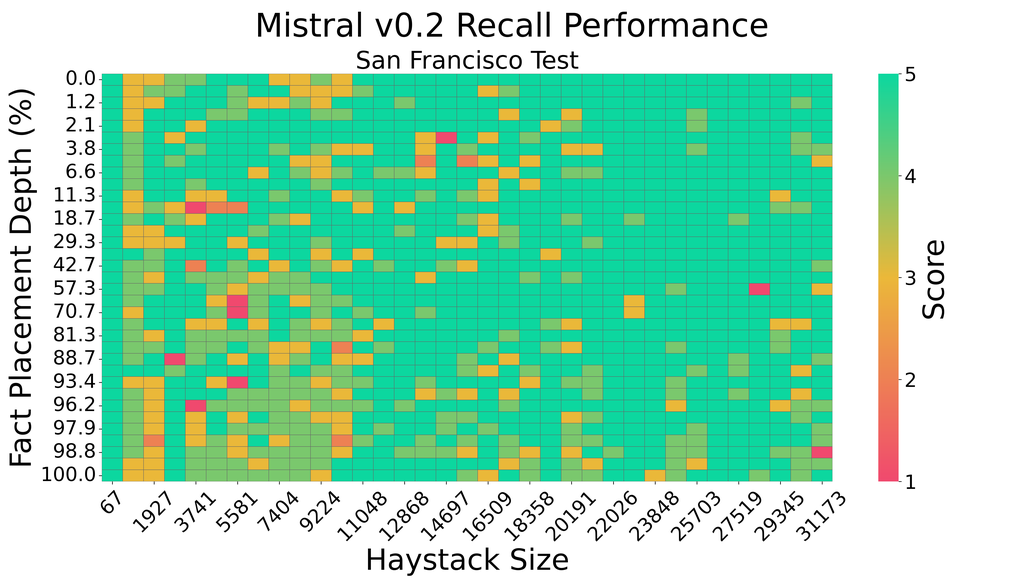}
\end{subfigure}
\end{figure}

\begin{figure}
\centering
\begin{subfigure}[b]{0.49\textwidth}
\includegraphics[width=\textwidth]{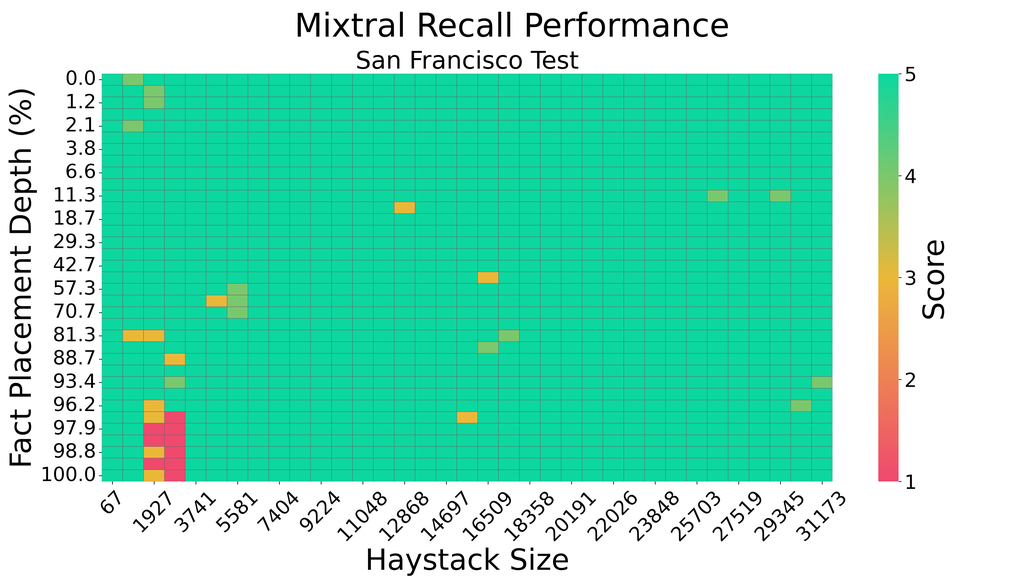}
\end{subfigure}  
\end{figure}

\clearpage
\newpage
\subsection{PistachioAI Heatmaps} 
\label{appendix:pistachio_ai_heatmaps}
LLM performance on the PistachioAI needle-in-a-haystack test.
\begin{figure}[H]
\centering
\begin{subfigure}[b]{0.49\textwidth}
\includegraphics[width=\textwidth]{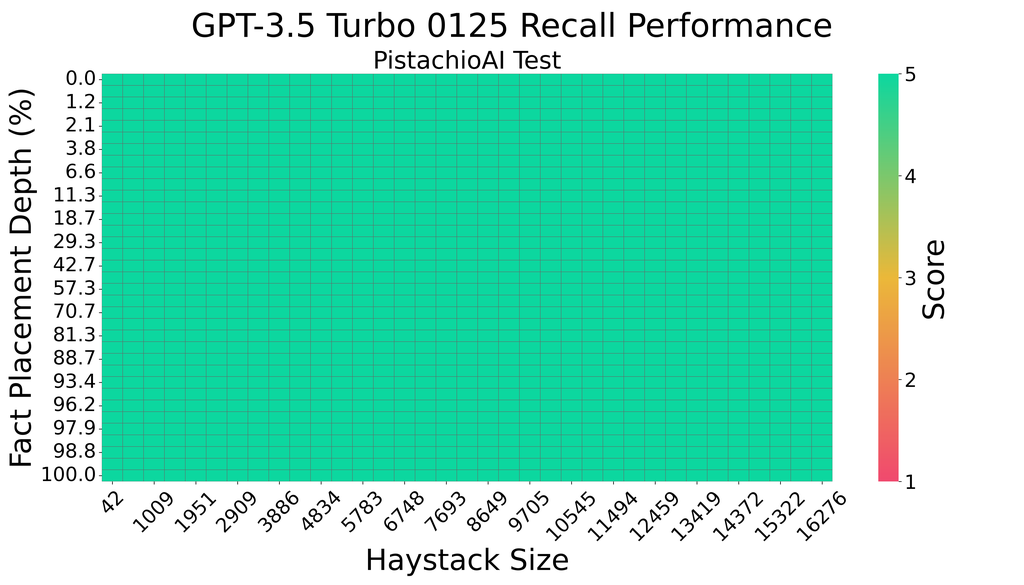}
\end{subfigure}
\hfill 
\begin{subfigure}[b]{0.49\textwidth}
\includegraphics[width=\textwidth]{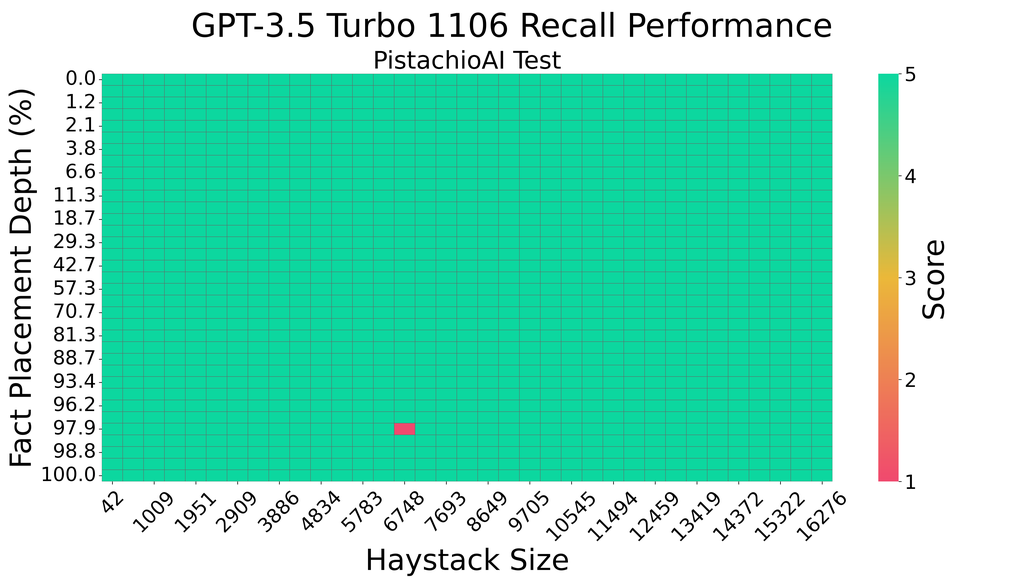}
\end{subfigure}
\end{figure}

\begin{figure}[H]
\centering
\begin{subfigure}[b]{0.49\textwidth}
\includegraphics[width=\textwidth]{pistachio_ai/gpt-4-0125-preview_pai_inline.png}
\end{subfigure}
\hfill
\begin{subfigure}[b]{0.49\textwidth}
\includegraphics[width=\textwidth]{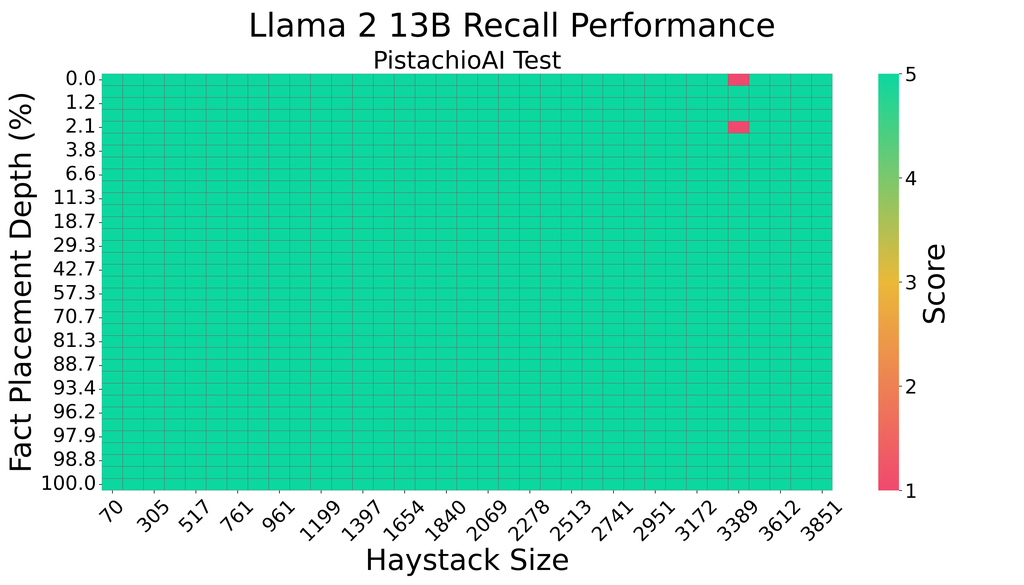}
\end{subfigure}
\end{figure}

\begin{figure}[H]
\centering
\begin{subfigure}[b]{0.49\textwidth}
\includegraphics[width=\textwidth]{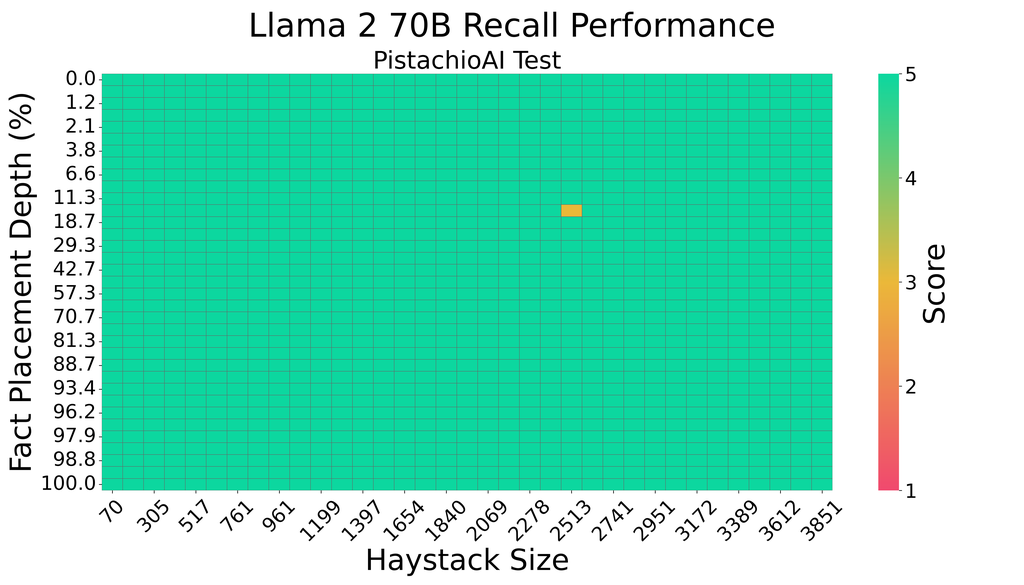}
\end{subfigure}
\hfill 
\begin{subfigure}[b]{0.49\textwidth}
\includegraphics[width=\textwidth]{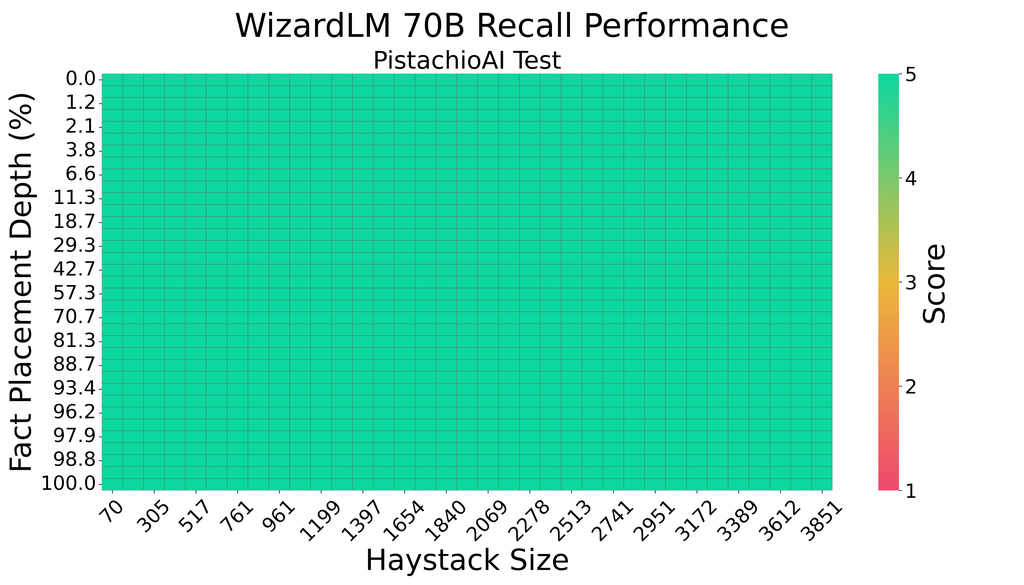}
\end{subfigure}
\end{figure}

\begin{figure}
\begin{subfigure}[b]{0.49\textwidth}
\includegraphics[width=\textwidth]{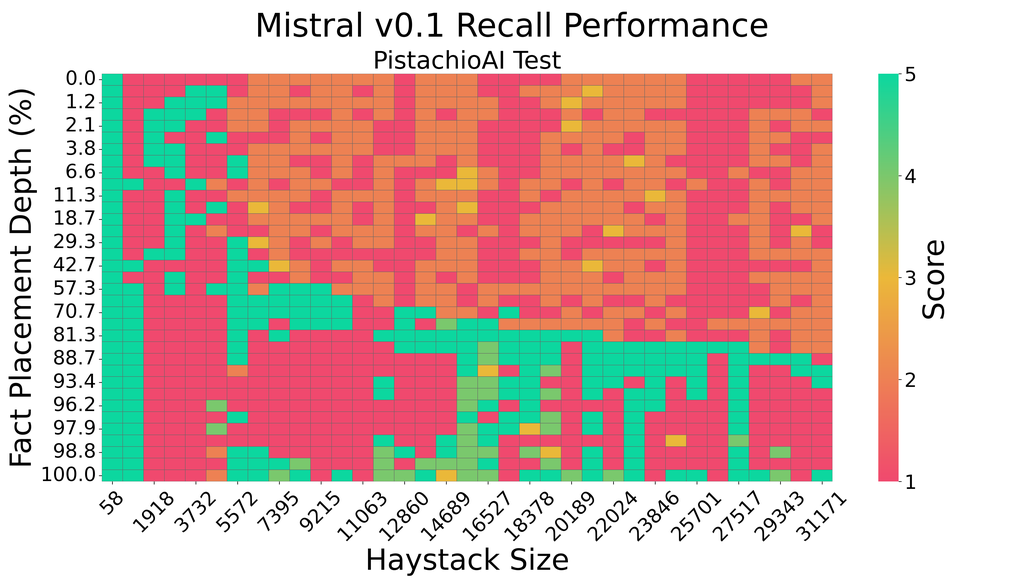}
\end{subfigure}
\hfill
\begin{subfigure}[b]{0.49\textwidth}
\includegraphics[width=\textwidth]{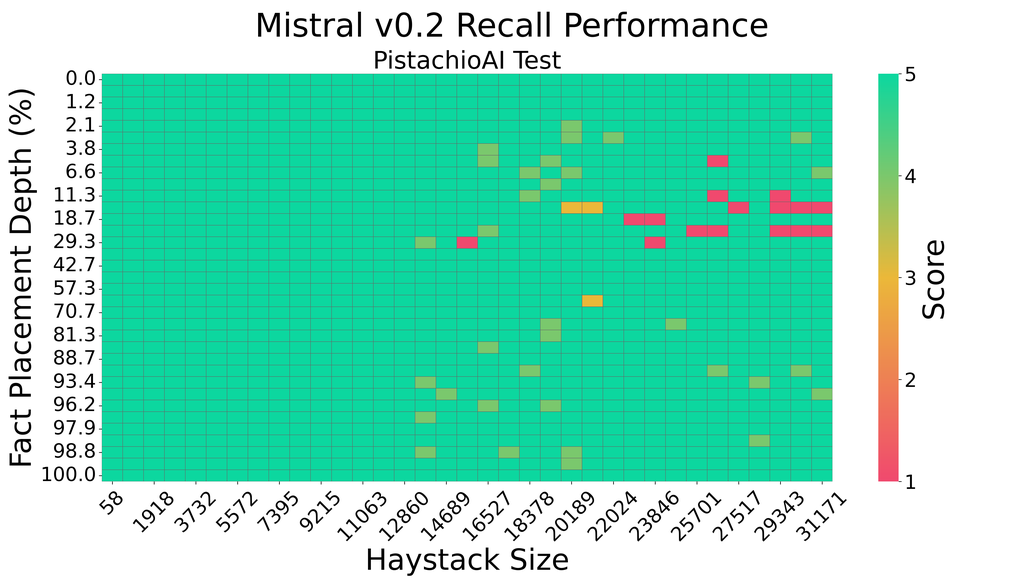}
\end{subfigure}
\end{figure}

\begin{figure}
\centering
\begin{subfigure}[b]{0.49\textwidth}
\includegraphics[width=\textwidth]{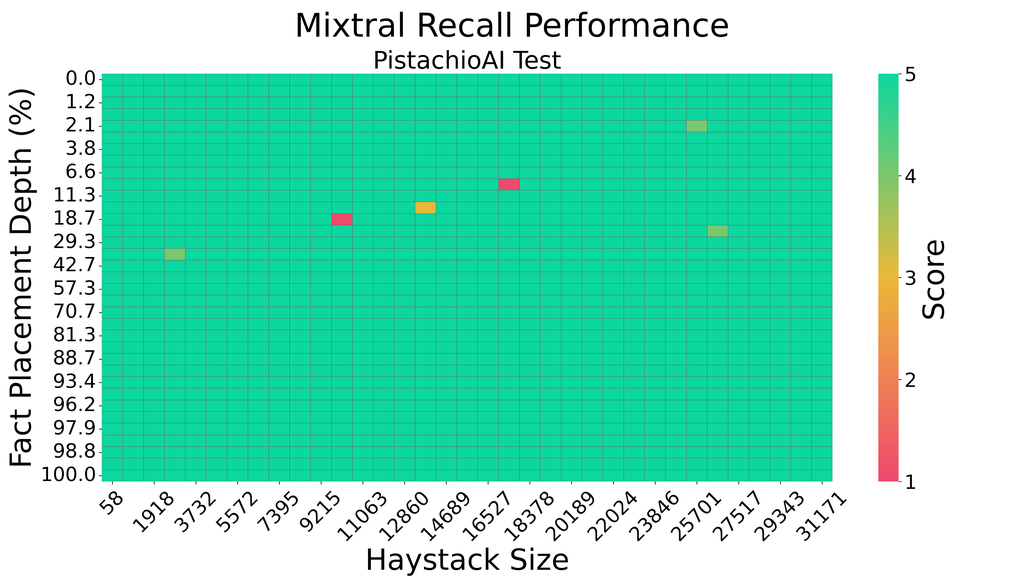}
\end{subfigure}  
\end{figure}

\end{document}